\documentclass[11pt]{article}

\usepackage[]{acl}

\usepackage{times}
\usepackage{latexsym}

\usepackage[T1]{fontenc}

\usepackage[utf8]{inputenc}

\usepackage{microtype}

\usepackage{inconsolata}

\usepackage{graphicx}
\usepackage{amsmath}
\usepackage{amssymb}
\usepackage{booktabs}
\usepackage{multirow}
\usepackage{tcolorbox}
\tcbuselibrary{skins}
\usepackage{pifont}
\usepackage{pgfplots}

\title{FlexComp: One Model for Every Ratio in Context Compression}

\author{Kaiyan Zhao\textsuperscript{1,2}, Zhongtao Miao\textsuperscript{1}, Akiko Aizawa\textsuperscript{2}, Yoshimasa Tsuruoka\textsuperscript{1}\\
        \textsuperscript{1}The University of Tokyo, \textsuperscript{2}National Institute of Informatics \\
        \texttt{\{kaiyan1006, miao, tsuruoka\}@logos.t.u-tokyo.ac.jp, aizawa@nii.ac.jp}}

\begin{document}
\maketitle
\begin{abstract}
Soft context compression condenses a context into a few memory tokens
that a frozen LLM consumes in place of the raw text, but existing
compressors fix the compression ratio at training and inference: each deployed
ratio requires a separately trained model, and the chosen ratio is
applied uniformly to all inputs, whose actual needs vary drastically.
We propose \textbf{FlexComp}, a method-agnostic framework that
decouples the ratio from both training and deployment: Matryoshka-style
training samples the memory budget $K$ per instance, turning one model
into an any-ratio compressor, and the budget is then chosen per input
by: (1) confidence-based cascade routing or (2) a lightweight learned
$K$~predictor. Across ICAE, 500xCompressor, and SAC on MRQA, a single
FlexComp model matches separately trained fixed-ratio specialists
with minimal degradation. Cascade routing preserves over 98\% of
the mildest ratio's accuracy at up to 266$\times$ average compression;
the $K$~predictor, in a single compression-decoding pass, reaches
158-236$\times$ within 0.7~F1\footnote{All F1 scores are reported on a 0-100 scale.} of the mildest ratio. At serving-scale batch sizes, the K predictor cuts context KV cache by 50\% and improves decoding throughput by 47\%.
\end{abstract}

\section{Introduction}

Large language models (LLMs) rarely answer from a question alone: in most applications, \emph{contexts}, such as retrieved passages, tool outputs, documents, or conversation history are needed for generation, besides the prompt~\citep{lewis2020retrieval, yuan-etal-2025-easytool, zhao2026ecomstagestagewiseorientationspecificbenchmarking}. Serving
these context tokens is costly, as they must be prefilled on every
query, and their Key-Value (KV) cache occupies memory for the
whole generation, often dominating the memory footprint of
inference \citep{liu2025kv}. \emph{Context
compression} offers an attractive remedy: a compressor condenses
the context into a small number of soft \emph{memory tokens}, and the LLM can run inference based on the compressed soft tokens instead of the raw context~\citep{ge2024icae, li-etal-2025-500xcompressor, zhao-etal-2025-position, liu2026sac, miao2026grcunifyingreasoningdrivengeneration}.

\begin{figure}[t]
    \centering
    \includegraphics[width=0.9\linewidth]{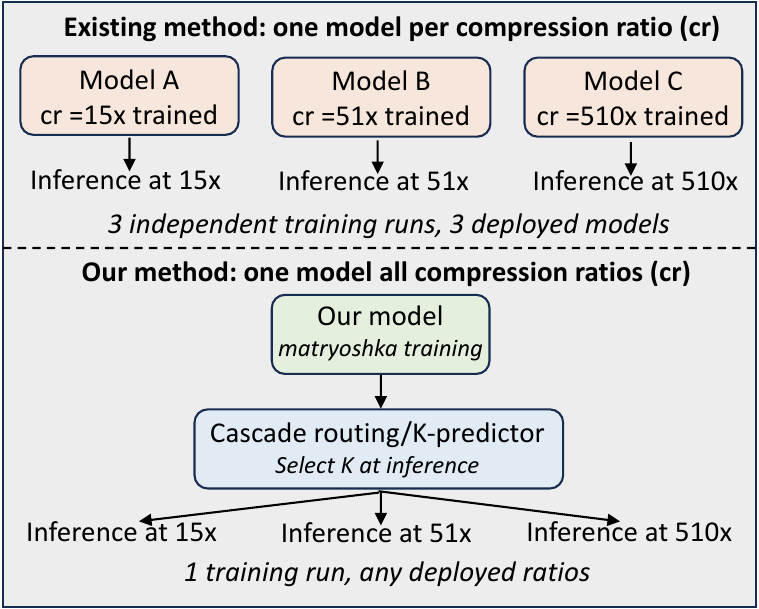}
    \caption{Fixed-ratio compression vs.\ FlexComp. \textbf{Top:}
    for existing soft-token compressors, each deployed compression ratio requires a separately trained model,
    and the model applies the fixed ratio uniformly to all
    inputs. \textbf{Bottom:} FlexComp trains a single model that supports any ratio, and selects the budget per input at inference time, via cascade routing or $K$ predictor.}
    \label{fig:teaser}
\end{figure}

Despite steady progress on compression quality, existing methods
share a structural limitation: \textbf{the compression ratio is
fixed at training and inference time}. As illustrated in
Figure~\ref{fig:teaser} (top), supporting multiple operating
modes, e.g., mild compression for difficult inputs and aggressive
compression when memory is scarce, requires training, storing, and
serving a separate model for every ratio. Moreover, whichever
ratio is deployed is applied \emph{uniformly} to all inputs,
regardless of how much each input actually needs.

\begin{figure}
    \centering
    \includegraphics[width=1.0\linewidth]{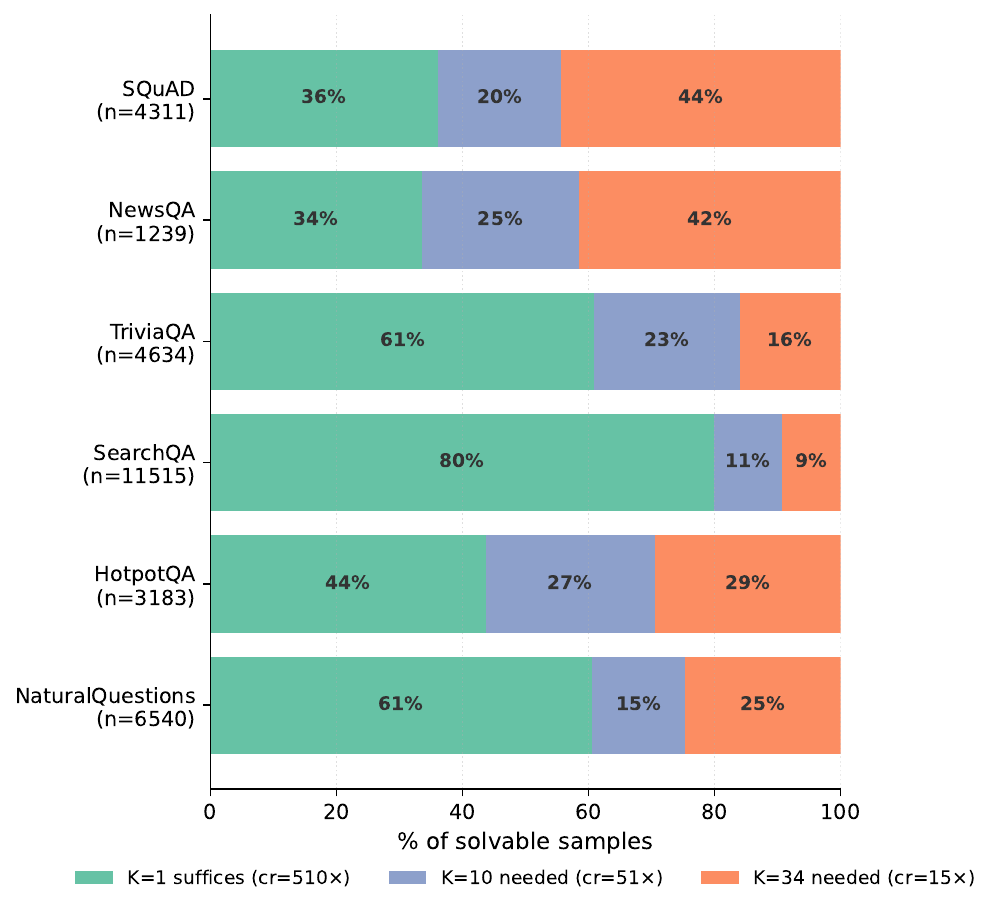}
     \caption{The budget each input needs varies drastically. For each subset, we show the distribution over solvable inputs of
    the \emph{minimal sufficient budget}: the smallest
    $K \in \{1, 10, 34\}$ (number of memory tokens) at which the SAC Matryoshka model
    answers correctly. A large fraction of inputs is already
    solvable at K{=}1 (510x compression), so any fixed ratio could be suboptimal.}
    \label{fig:motivation}
\end{figure}

This uniformity is wasteful, because different inputs need very
different budgets. Figure~\ref{fig:motivation} shows that for the
in-domain subsets in MRQA~\citep{fisch-etal-2019-mrqa}, most inputs are already solved with a single memory
token (soft token length $K=1$)\footnote{Following existing work, contexts are segmented into
chunks of 510 tokens; $K{=}1$ thus means one soft memory token per
510 raw tokens, i.e., a compression ratio of 510x.}, yet a non-trivial minority is only solved with far larger budgets. A fixed $K$ therefore loses on both ends: it wastes
tokens on the many easy inputs and starves the few hard ones.
Worse, the best fixed choice differs from domain to domain, so no
single ratio serves all deployments well.

To this end, we propose \textbf{FlexComp}, a framework that decouples the
compression ratio from both training and deployment
(Figure~\ref{fig:teaser}, bottom). FlexComp has two components.
First, we introduce \emph{Matryoshka-style training} for context
compression, inspired by matryoshka representation learning
\citep{kusupati2022matryoshka}. Specifically, during training, the soft prompt length $K$ is sampled
per instance, so that a \emph{single} model learns to encode the
context into $K$ memory tokens for any budget in the supported range.
One model thus replaces an entire family of
fixed-ratio compressors, with no architectural changes to the
underlying method. Second, since the budget is now a free
inference-time variable, we introduce two complementary strategies
for choosing it per input: \emph{cascade routing}, which starts from
the most aggressive budget and re-encodes at a larger one only when
the decoder is uncertain, and a lightweight \emph{$K$ predictor},
which commits to a budget \emph{before} compression and thus answers
in a single compression-decoding pass. FlexComp is method-agnostic;
we instantiate it on three representative compressors:
ICAE~\citep{ge2024icae},
500xCompressor~\citep{li-etal-2025-500xcompressor}, and
SAC~\citep{liu2026sac}.

Across in-domain (ID) and out-of-domain (OOD) question answering tasks on
MRQA~\citep{fisch-etal-2019-mrqa}, a single Matryoshka-trained model
matches separately trained fixed-ratio specialists at each native
ratio (within 1.3 F1 in the worst case, and often better on OOD),
cutting the number of trained models from $n$ to one. On top of the
same model, adaptive budget selection improves the
accuracy-compression trade-off over any fixed ratio: cascade routing
retains over 98\% of the mildest ratio's accuracy while raising the
average compression from 15$\times$ to up to 266$\times$, and the
$K$~predictor reaches 158--236$\times$ in a single
compression-decoding pass, at a cost of at most 0.7~F1. These
savings are not merely nominal: at serving-scale batch sizes, the
predictor cuts context KV cache by 50\% and improves aggregate
decoding throughput by 47\% relative to always running at the
mildest ratio.

In summary, our contributions are as follows:
\begin{itemize}
\item We show that Matryoshka training turns a fixed-ratio context
compressor into a single \emph{any-ratio} compressor, validated on
three distinct methods with negligible degradation.

\item We introduce two per-input budget-selection strategies for
soft context compression: confidence-driven cascade routing and the $K$ predictor that commits to a budget before
compression; for the predictor, we further show these savings translate into gains in KV memory and decoding throughput.

\item We systematically analyze the accuracy-compression trade-off, and find it non-monotonic at the instance
level: on a non-trivial fraction of inputs, larger budgets preserve
surface details that may mislead the decoder, and a single memory token
answers correctly where the full budget fails.
\end{itemize}
\section{Related Work}
\label{sec:related}
Context compression methods fall into two families: \emph{soft}
compression, which encodes the context into continuous memory
tokens, and \emph{hard}
compression, which shortens the prompt in token space by filtering
or rewriting the discrete text. Our work builds on the former.
 
\paragraph{Soft context compression.}
A growing line of work compresses a context into a small number of continuous memory tokens consumed by a frozen LLM, including gist tokens \citep{mu2023gist}, AutoCompressors \citep{chevalier2023adapting}, ICAE \citep{ge2024icae}, 500xCompressor \citep{li-etal-2025-500xcompressor}, EPL~\citep{zhao-etal-2025-position} and SAC \citep{liu2026sac}. These methods differ in soft token acquisition and supervision signals, but the core idea, representing contexts through soft tokens, remains consistent. Moreover, they share the same limitation that the compression ratio is uniformly a training-time constant: each deployed ratio requires its own trained model, and all inputs receive the same budget. FlexComp targets exactly on this axis and is complementary to all base architectures.
 
\paragraph{Hard context compression.}
An orthogonal family shortens the prompt in token space, by filtering or summarizing the discrete text, e.g., Selective Context \citep{li-etal-2023-compressing-selective-context}, the LLMLingua series \citep{jiang-etal-2024-longllmlingua, pan-etal-2024-llmlingua}, and RECOMP \citep{xu2024recomp}. Hard compression preserves interpretability, but achievable ratios are typically modest compared to soft methods. As with soft compression methods, how much to compress a given input remains an open question, which our budget-selection strategies are designed to address.
 
\paragraph{Nested and budget-conditioned representations.}
Matryoshka representation learning trains embeddings whose prefixes remain valid at multiple dimensionalities \citep{kusupati2022matryoshka}, a principle that can be extended to model widths and other capacity axes \citep{devvrit2024matformer}. We transfer the one-model-many-budgets principle to context compression, with a structural difference: our budgets are realized by conditioning the encoder on the memory slot count rather than by truncating a fixed representation (\S\ref{sec:matryoshka}).
 
\paragraph{Adaptive computation and cascades.}
Allocating inference cost per input has a long history, from early exit \citep{miao2024efficientinferenceframeworkearlyexit} to LLM cascades that escalate from cheap to expensive models based on confidence \citep{yue2024large}. Our cascade applies this recipe within a single model, escalating over compression budgets rather than over models, and our $K$ predictor mirrors router-based approaches that commit to a compute level before execution \citep{stripelis-etal-2024-tensoropera}. Relatedly, KV cache eviction and quantization \citep{zhang2023h2oheavyhitteroracleefficient, 3737916.3738638} reduce memory after the full context has been prefilled; soft-token compression avoids materializing the full-context cache in the first place.
\section{Method}
\label{sec:method}
\begin{figure*}
    \centering
    \includegraphics[width=1.0\linewidth]{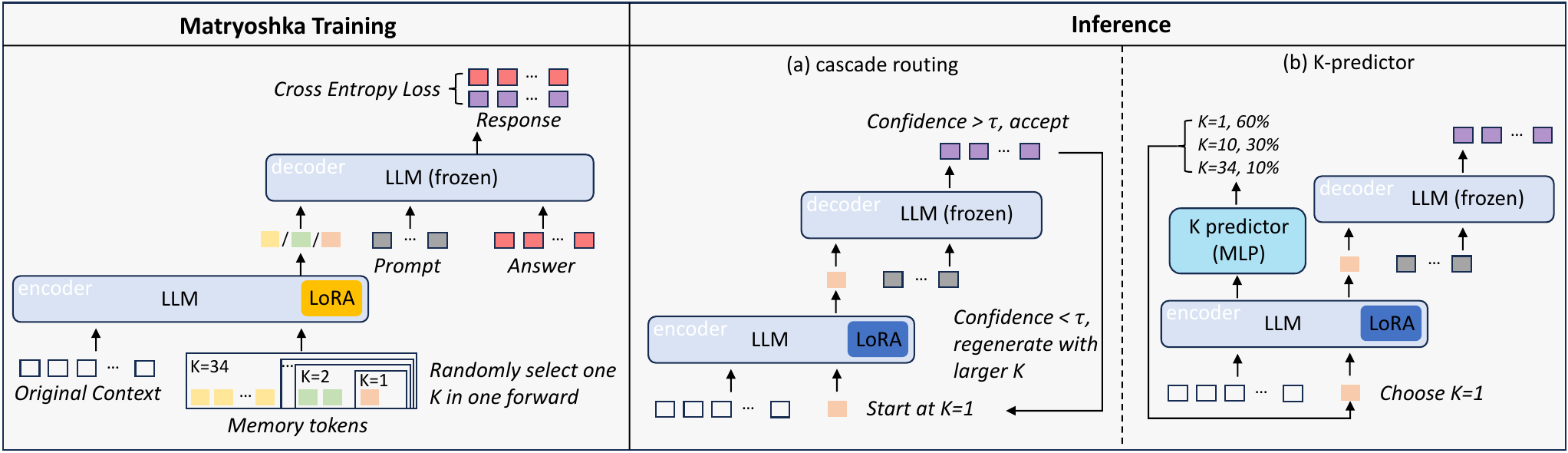}
    \caption{Overview of FlexComp. \textbf{Left:} Matryoshka training in fine-tuning. For
each instance, a budget $K$ is randomly sampled. The LoRA-adapted encoder compresses the context into
$K$ memory tokens, and the frozen decoder is supervised to generate the answer based on the memory tokens and prompt. \textbf{Right:}
two inference-time budget-selection strategies exploiting the
resulting freedom. \textbf{(a)}~Cascade routing starts at the most
aggressive budget and re-encodes at the next larger one only when the
decoder's confidence falls below the threshold~$\tau$.
\textbf{(b)}~The $K$~predictor is a lightweight MLP head that maps
budget-free encoder features to a distribution over $\mathcal{K}$ and
commits to a budget before compression, answering in a single
compression-decoding pass.}
    \label{fig:overview_method}
\end{figure*}

Figure~\ref{fig:overview_method} gives an overview of FlexComp. We first fix notation and the class of compressors we build on (\S\ref{sec:prelim}), then describe Matryoshka training, which converts a fixed-ratio compressor into a single any-budget compressor (\S\ref{sec:matryoshka}), and finally two inference-time strategies for choosing the budget per input: confidence-based cascade routing (\S\ref{sec:cascade}) and a learned $K$ predictor (\S\ref{sec:kpred}).
 
\subsection{Preliminaries: Soft-Token Context Compression}
\label{sec:prelim}
 
We first consider an encoder LLM $E_\phi$, that compresses a context $c$ of $L$ tokens into
$K \ll L$ continuous \emph{memory tokens}:
$\mathbf{m}_{1:K} = E_\phi(c; K)$, which a frozen decoder LLM $D$
then consumes in place of $c$ to answer a query $q$, i.e.,
$\hat{a} = D(\mathbf{m}_{1:K}, q)$. In practice, the encoder
$E_\phi$ is a LoRA-adapted copy of
$D$, that processes the context together with $K$ trainable memory tokens. The decoder consumes the representations of these memory tokens after the encoder as a drop-in prefix, so the context's footprint in $D$'s KV
cache shrinks from $L$ to $K$ entries. This formulation covers
ICAE~\citep{ge2024icae},
500xCompressor~\citep{li-etal-2025-500xcompressor}, and
SAC~\citep{liu2026sac}, which differ in where the memory tokens are placed
and how the encoder is supervised, but share the same interface: the
number of memory slots $K$ is fixed before training, and the
resulting model supports only that budget. Throughout, we refer to
$K$ as the \emph{budget}; with $|c| = L$, the corresponding
compression ratio is $L/K$.

\paragraph{Training objectives.}
Existing compressors are trained in two stages, both at a fixed $K$.
A \emph{pretraining} stage teaches the encoder to pack generic text
into memory tokens: ICAE and 500xCompressor supervise the decoder to
regenerate or continually generate the context from $\mathbf{m}_{1:K}$ (autoencoding or text continuation objective), while SAC directly optimizes language modeling
conditioned on its semantic anchors. Abstractly,
\begin{equation}
\mathcal{L}_{\mathrm{PT}}(\phi)=
\mathbb{E}_{c}\!\left[\ell\!\left(D(E_\phi(c;K)),\, t(c)\right)\right],
\label{eq:pt}
\end{equation}
where the target $t(c)$ is the context itself or its continuation. A
subsequent \emph{fine-tuning} stage adapts the model to downstream
use, supervising the answer given the compressed context:
\begin{equation}
\mathcal{L}_{\mathrm{SFT}}(\phi)=
\mathbb{E}_{(c,q,a)}\!\left[\ell\!\left(D(E_\phi(c;K), q),\, a\right)\right].
\label{eq:sft}
\end{equation}
FlexComp keeps both objectives untouched and changes only how $K$ is
set: Matryoshka sampling (Eq.~\ref{eq:matryoshka}) wraps
$\mathcal{L}_{\mathrm{SFT}}$, while pretraining deliberately remains
at a fixed budget, a choice we justify empirically in
§\ref{sec:ana_pretrain}.

\subsection{Matryoshka Training}
\label{sec:matryoshka}
 
FlexComp makes the budget a \emph{conditioning variable} rather than an architectural constant (Figure~\ref{fig:overview_method}, left). Let $\mathcal{K} = \{K_1 < K_2 < \dots < K_n\}$ be the set of supported budgets. During training, for each instance we sample a budget $K \sim p(\mathcal{K})$, allocate $K$ memory slots, and optimize the same objective as the underlying compressor:
\begin{equation}
\mathcal{L}(\phi) = \mathbb{E}_{(c,q,a)}\, \mathbb{E}_{K \sim p(\mathcal{K})} \big[ \ell\big(D(E_\phi(c;K), q), a\big) \big],
\label{eq:matryoshka}
\end{equation}
where $\ell$ is the token-level cross-entropy of the base method and $D$ remains frozen. The single set of weights $\phi$ is thus trained to produce, for every $K \in \mathcal{K}$, a representation \emph{specialized to that budget}: the encoder sees how many slots are available and distributes information accordingly. We emphasize that, unlike Matryoshka representation learning for embeddings \citep{kusupati2022matryoshka}, the budgets are not realized by truncating a single encoding, as $E_\phi(c; K_1)$ is a different forward pass from the first $K_1$ tokens of $E_\phi(c; K_2)$, but the training principle is shared: one model, nested budgets, joint supervision. 
 
Matryoshka training requires no architectural changes to the base compressor and adds no inference-time overhead relative to a fixed-ratio model evaluated at the same $K$. It collapses $n$ training runs and $n$ deployed models into one, and exposes the full budget range $\mathcal{K}$ at inference time.
 
\subsection{Inference I: Cascade Routing}
\label{sec:cascade}
 
The first strategy for choosing the appropriate budget per input requires no additional training (Figure~\ref{fig:overview_method}, Inference (a)). Given an input, we start from the most aggressive budget $K_1$ and iteratively escalate. At stage $i$, we encode $\mathbf{m}_{1:K_i} = E_\phi(c; K_i)$, decode a candidate answer, and compute a confidence score $s_i$ from the decoder's output log-probabilities, defined as the length-normalized log-probability
of the decoded answer:
\begin{equation}
    s_i = \tfrac{1}{|\hat{a}_i|}\sum_{t} \log p_\theta(\hat{a}_{i,t}).
\end{equation}

This score is a reliable proxy for answer quality: it correlates
strongly with token-level F1 at every budget
(Spearman's $\rho$ up to 0.58; §\ref{sec:confidence-analysis}).
If $s_i \geq \tau_i$, the candidate is accepted; otherwise we discard it and re-encode at the next budget $K_{i+1}$. The final budget $K_n$ always accepts, so the procedure terminates. 
The thresholds $\tau_1, \dots, \tau_{n-1}$ are calibrated once on held-out data by sweeping the accuracy-average-budget trade-off. Different threshold settings trace out a controllable operating curve between the most aggressive and mildest budgets. 
 
\subsection{Inference II: Learned $K$ Prediction}
\label{sec:kpred}
 
The cascade decides \emph{after} decoding; our second strategy decides \emph{before} compression (Figure~\ref{fig:overview_method}, Inference (b)), committing to a single $K$ per input.
 
\paragraph{Offline label construction.}
We derive supervision directly from the Matryoshka model itself. For
each training instance $(c,q,a)$, we decode at every $K \in
\mathcal{K}$ and score the output against the gold answer with
Rouge-1 F1, yielding $F_K$. The label is the smallest budget whose
score is within a tolerance $\varepsilon$ of the best achievable
one,
\begin{equation}
K^{*} \;=\; \min\bigl\{\,K \in \mathcal{K} \;:\; F_K \,\ge\,
F_{K_n} - \varepsilon \,\bigr\},
\label{eq:label}
\end{equation}
with $\varepsilon{=}0.1$. Instances that remain low-quality even at
the full budget ($F_{K_n} < 0.5$) are excluded from predictor
training. This turns budget selection into
an $n$-way classification problem requiring no human annotation.
 
\paragraph{Class balancing.}
The label distribution is heavily skewed: most instances are solvable at $K_1$ (Figure~\ref{fig:motivation}). Trained on the raw distribution, the predictor collapses to a shortcut of predicting the majority class. We therefore resample the training set to a balanced $1{:}1{:}\dots{:}1$ ratio across the $n$ classes, which we found necessary and sufficient to prevent this collapse (\S\ref{sec:ana_predictor}). 
 
\paragraph{Predictor.}
The predictor operates on a budget-independent representation of the
context: the context chunks are passed through the encoder
\emph{without} allocating any memory slots, and the final-layer
hidden states are mean-pooled over all context tokens into a single
vector $\mathbf{h} \in \mathbb{R}^{d}$. A two-layer MLP
($d \rightarrow d/2$, ReLU, $d/2 \rightarrow n$) maps $\mathbf{h}$
to a distribution over $\mathcal{K}$. Only the head is trained with cross-entropy on the labels of Eq.~\ref{eq:label} on the balanced dataset. Because the features are computed from
the context alone, the predicted budget is query-agnostic: it can be
computed once per document, offline, and cached alongside the
compressed representation. At inference, budget selection costs one
budget-free encoder forward plus the (negligible) head; the input is
then encoded once at the predicted $\hat{K}$ and decoded once.
Compared with the cascade, the predictor forgoes the decoder's own
confidence signal but replaces up to $n$ encode-decode rounds with
this fixed cost, and the two occupy complementary points on the
cost-accuracy spectrum.
\section{Experiments}

\begin{table*}[t]
\centering
\small
\resizebox{0.65\textwidth}{!}{
\begin{tabular}{ll cc cc cc}
\toprule
& & \multicolumn{2}{c}{\text{\shortstack{Fixed-Ratio\\(3 models per method)}}} & \multicolumn{2}{c}{\text{\shortstack{Matryoshka\\(1 model, \textbf{ours})}}} & \multicolumn{2}{c}{\textbf{\shortstack{$\Delta$\\(Matr.\ $-$ Fixed)}}} \\
\cmidrule(lr){3-4} \cmidrule(lr){5-6} \cmidrule(lr){7-8}
\textbf{Base} & \textbf{$K$ (ratio)} & \textbf{ID} & \textbf{OOD} & \textbf{ID} & \textbf{OOD} & \textbf{ID} & \textbf{OOD} \\
\midrule
\multirow{3}{*}{ICAE}
 & 34 (15$\times$)  & 44.70 & 30.11 & 43.46 & 30.62 & $-1.24$ & $+0.51$ \\ 
 & 10 (51$\times$)  & 41.17 & 28.05 & 41.82 & 29.10 & $+0.65$ & $+1.05$ \\
 & 1 (510$\times$)  & 36.87 & 26.85 & 35.88 & 26.19 & $-0.99$ & $-0.66$ \\
\midrule
\multirow{3}{*}{500x}
 & 34 (15$\times$)  & 49.13 & 35.22 & 48.80 & 34.62 & $-0.33$ & $-0.60$ \\
 & 10 (51$\times$)  & 43.07 & 30.72 & 42.79 & 30.68 & $-0.28$ & $-0.04$ \\
 & 1 (510$\times$)  & 34.57 & 25.18 & 38.29 & 28.30 & $+3.72$ & $+3.12$ \\
\midrule
\multirow{3}{*}{SAC}
 & 34 (15$\times$)  & 54.89 & 40.07 & 53.82 & 40.10 & $-1.07$ & $+0.03$ \\
 & 10 (51$\times$)  & 46.32 & 32.29 & 47.09 & 33.22 & $+0.77$ & $+0.93$ \\
 & 1 (510$\times$)  & 39.05 & 26.93 & 39.63 & 28.13 & $+0.58$ & $+1.20$ \\
\bottomrule
\end{tabular}}
\caption{F1 of separately trained fixed-ratio compressors vs.\ a single Matryoshka-trained model per method, averaged over in-domain (ID) and out-of-domain (OOD) tasks. $K{=}34/10/1$ correspond to $15\times/51\times/510\times$ compression.}
\label{tab:matryoshka_compact}
\end{table*}

\subsection{Experimental Setup}
\label{sec:setup}

\paragraph{Compressors and budgets.}
We instantiate FlexComp on three soft context compressors: ICAE \citep{ge2024icae}, 500xCompressor \citep{li-etal-2025-500xcompressor}, and SAC \citep{liu2026sac}, keeping each method's architecture, training objective, and hyperparameters unchanged; the only modification is budget sampling (\S\ref{sec:matryoshka}). All compressors use Llama-3.2-1B\footnote{\url{https://huggingface.co/meta-llama/Llama-3.2-1B}}~\citep{grattafiori2024llama3herdmodels} as both the encoder and decoder, with the encoder equipped with LoRA~\citep{hu2022lora} adapters ($r$=128, $\alpha$=256). 
We split the contexts into chunks with length 510, and we use budgets $\mathcal{K} = \{1, 10, 34\}$, i.e., compression ratios of $510\times$, $51\times$, and $15\times$. During Matryoshka training, $K$ is sampled uniformly from $\mathcal{K}$ per instance. 

\paragraph{Tasks and metrics.}
We follow the evaluation protocol of existing methods and evaluate our models on MRQA~\citep{fisch-etal-2019-mrqa}, including six in-domain (ID) tasks (SQuAD, NewsQA, TriviaQA, SearchQA, HotpotQA, NaturalQuestions) and six out-of-domain (OOD) tasks (BioASQ, DROP, DuoRC, RACE, RelationExtraction, TextbookQA).
The compressor is continually pretrained on the SlimPajama-6B dataset~\citep{cerebras2023slimpajama}, and finetuned on the MRQA training set. We report token-level F1 averaged within each group.
 
\paragraph{Baselines.}
For each method, we compare the single Matryoshka-trained model against three \emph{fixed-ratio specialists}: the same architecture trained conventionally at each $K \in \mathcal{K}$, following the original recipes. All fixed-ratio baselines are trained by us under identical data and compute budgets for a controlled comparison. 
 
\paragraph{Budget selection.}
For cascade routing, thresholds are calibrated
once on held-out data by sweeping the accuracy--average-budget
trade-off; $K_n{=}34$ always accepts, so each operating point is a
pair $(\tau_{K=1}, \tau_{K=10})$ over the length-normalized answer
log-probability. We report four points of increasing permissiveness:
\emph{conservative} $(-0.20, -0.30)$, \emph{balanced}
$(-0.20, -0.40)$, \emph{aggressive} $(-0.30, -0.40)$, and
\emph{high-cr} $(-0.40, -0.50)$; a lower (more negative) threshold
accepts lower-confidence candidates earlier and thus yields a higher
average compression ratio. The $K$ predictor head adds 2.1M
parameters for Llama-3.2-1B ($d{=}2048$; $0.17\%$ of the encoder).
 
\paragraph{Implementation Details.}
All training runs on 8 H800 GPUs with a per-GPU batch size of 16.
Compressor pretraining and fine-tuning are each conducted for 20,000
steps with AdamW (lr $5{\times}10^{-4}$). For the $K$ predictor,
labeling the MRQA training split (Eq.~5) yields 183K samples
(60.85\% $K{=}1$, 18.00\% $K{=}10$, 21.15\% $K{=}34$); we downsample
to a 1:1:1 balance ($\sim$99K) to prevent the shortcut collapse
analyzed in §\ref{sec:ana_predictor}, and train for 2 epochs with the same optimizer,
updating only the predictor head while the compressor stays frozen. For pre-training, we fix $K=34$ (15$\times$ compression).

\subsection{One Model Matches Ratio-Specific Models}
\label{sec:exp_matryoshka}
Table~\ref{tab:matryoshka_compact} compares each fixed-ratio
specialist with the single Matryoshka-trained model evaluated at the
same budget. Across all three architectures and all three ratios, the
gap is small.
Replacing three trained models with one costs at most about one F1
point, and in four of nine settings the single model is in fact
better. On OOD tasks, the picture tilts further in favor of
Matryoshka training: seven of nine deltas are positive, and the only
notable losses occur for ICAE at the extreme $510\times$ ratio. The
clearest win is 500xCompressor at $510\times$, where the shared model
exceeds its specialist by $+3.72$ ID and $+3.12$ OOD F1. We
conjecture that jointly training across budgets regularizes the
extreme $K{=}1$ case, which a dedicated specialist must otherwise
learn in isolation. Overall, turning the budget into a free
inference-time variable costs essentially no accuracy, while cutting
the number of trained and deployed models from $n$ to one.

\begin{figure*}[t]
    \centering
    \includegraphics[width=1.0\linewidth]{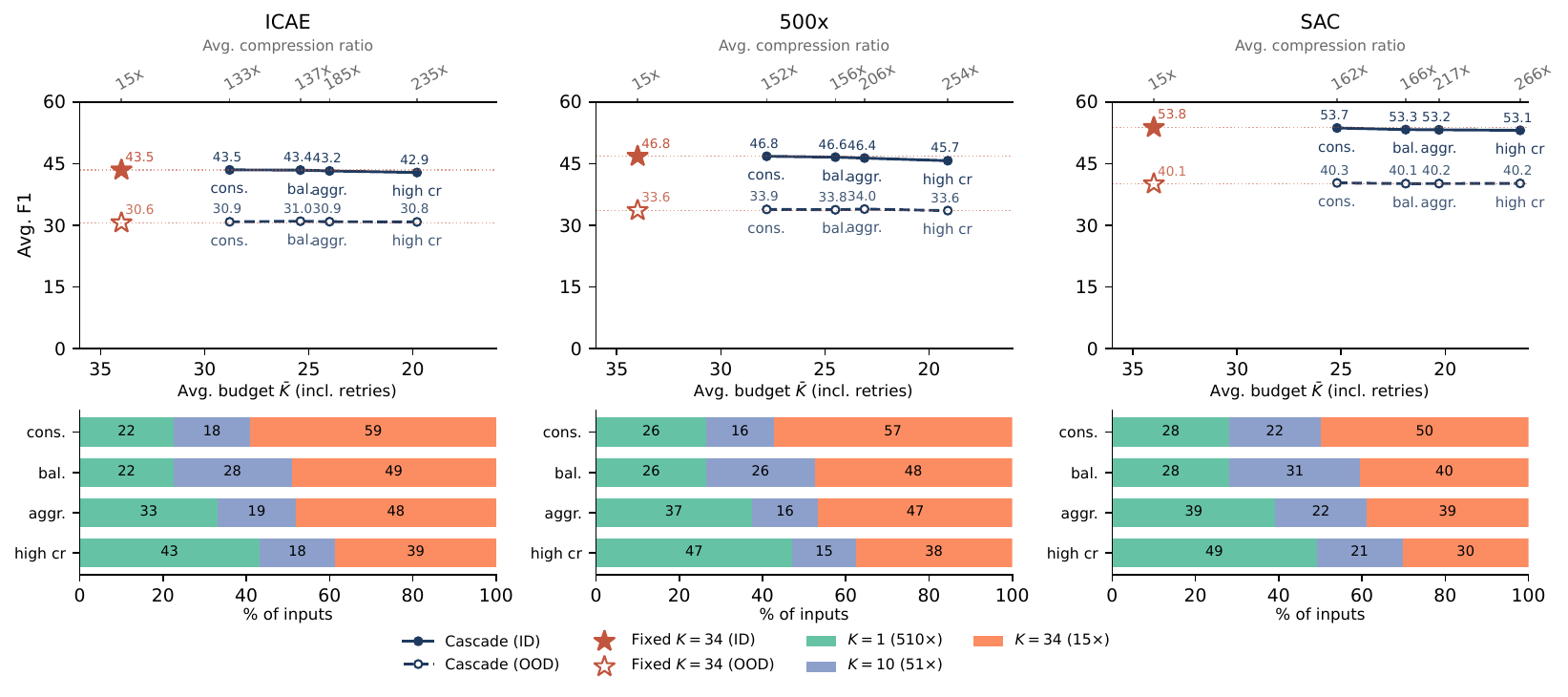}
    \caption{Cascade routing results on ICAE, 500xCompressor, and SAC. \textbf{Top:} average F1 (ID: solid, OOD: dashed, F1 reported above each mark) versus the average memory budget $\bar{K}$, where $\bar{K}$ includes retry costs after escalation; the top axis reports the corresponding measured average compression ratio. Each curve traces four cascade configuration thresholds, ordered from conservative (cons.) to most aggressive (high~cr), and stars mark the fixed $K{=}34$ baseline of the same Matryoshka model. \textbf{Bottom:} realized routing distribution at each configuration.}
    \label{fig:cascade_main}
\end{figure*}
\subsection{Cascade Routing}
Figure~\ref{fig:cascade_main} shows what happens when the cascade chooses the budget at inference time. The pattern is the same for all three methods: as the thresholds get more permissive, the average budget drops sharply while accuracy barely moves. On ICAE, the most aggressive operating point uses an average of $19.8$ tokens per input, an average compression ratio of $235\times$, yet stays within $0.6$ ID F1 of always running at $K{=}34$, and OOD accuracy is slightly \emph{above} the fixed baseline at every operating point. 
The routing distributions (bottom row) explain why this works: even at conservative thresholds, over 20\% of inputs are confidently answered at $K{=}1$, matching the oracle picture in Figure~\ref{fig:motivation}. The cascade thus turns the thresholds into a practical control knob: one deployed model, and a compression-accuracy trade-off that can be tuned without retraining.

\subsection{$K$ Predictor}
\label{sec:exp_kpred}
\begin{table*}[ht]
\centering
\small
\setlength{\tabcolsep}{6pt}
\resizebox{0.85\textwidth}{!}{
\begin{tabular}{lcccc ccc}
\toprule
& & & & & \multicolumn{3}{c}{\textbf{Predicted budget (\% of inputs)}} \\
\cmidrule(lr){6-8}
\textbf{Model} & \textbf{ID F1} & \textbf{OOD F1} & \textbf{Avg.\ cr} & \textbf{Avg.\ $\bar{K}$}
& \textbf{$K{=}1$ ($510\times$)} & \textbf{$K{=}10$ ($51\times$)} & \textbf{$K{=}34$ ($15\times$)} \\
\midrule
ICAE ($K{=}34$)       & 43.46 & 30.62 & $15\times$  & 34.0 & --    & --    & 100.0 \\
\; + $K$ predictor    & 43.29 & 30.61 & $157.99\times$ & 19.6 & 27.2 & 22.7 & 50.1 \\
\; + random routing   & 40.02 & 28.03 & $157.99\times$ & 19.6 & 27.2 & 22.7 & 50.1 \\
\midrule
500x ($K{=}34$)       & 48.80 & 34.62 & $15\times$  & 34.0 & --    & --    & 100.0 \\
\; + $K$ predictor    & 48.28 & 34.20 & $236.08\times$ & 16.9 & 43.9 & 11.0 & 45.1 \\
\; + random routing   & 42.61 & 30.96 & $236.08\times$ & 16.9 & 43.9 & 11.0 & 45.1 \\
\midrule
SAC ($K{=}34$)        & 53.82 & 40.10 & $15\times$  & 34.0 & --    & --    & 100.0 \\
\; + $K$ predictor    & 53.14 & 39.95 & $158.96\times$ & 19.0 & 27.3 & 25.0 & 47.7 \\
\; + random routing   & 48.25 & 35.14 & $158.96\times$ & 19.0 & 27.3 & 25.0 & 47.7 \\
\bottomrule
\end{tabular}}
\caption{Per-input budget selection with the $K$ predictor. Each pair compares the Matryoshka model run uniformly at its mildest budget ($K{=}34$, $15\times$) against the same model with the predictor choosing $K \in \{1, 10, 34\}$ per input before encoding. ID/OOD: average F1 over in-/out-of-domain tasks; Avg.\ cr / $\bar{K}$: measured average compression ratio and memory budget; right block: distribution of predicted budgets. \emph{Random routing} assigns budgets with the same per-budget proportions but ignores the input, matching the predictor's average budget.}
\label{tab:kpredictor}
\end{table*}
The $K$~predictor decides the budget for each input before compression and pays for a single compression-decoding pass. Table~\ref{tab:kpredictor} shows this
cheaper strategy works well: relative to always running at the
mildest budget, the predictor roughly halves the average budget and
raises the compression ratio to 158--236$\times$, while ID and OOD
accuracy both drop by at most 0.7~F1.

The random routing baseline isolates where this benefit comes from.
It assigns the same per-budget proportions as the predictor but
ignores the input, so it matches the predictor's average budget
exactly; the predictor still outperforms it by about 5~F1 on both ID
and OOD. The gain is therefore not an artifact of the budget mix
being favorable on average, it comes from routing the \emph{right}
inputs to the smaller budgets, i.e., the predictor identifies which
inputs tolerate aggressive compression rather than merely
compressing more.

\subsection{Which Strategy to Deploy}

Together, the two strategies serve complementary deployment needs. When a corpus is
compressed once and queried repeatedly,
the cascade's retry cost disappears while its
advantage remains: it adapts to each question with iteratively decided ratio, thus offering a tunable, retraining-free trade-off curve for this scenario.
When each context is compressed on the fly for a single query, the predictor gives a single-pass operating point cheap
enough for latency-sensitive serving.
Its uniform per-request cost is also what makes the budget-grouped
batching of §\ref{sec:throughput} possible.

\section{Analysis}
\label{sec:analysis}

\subsection{Matryoshka in Pretraining Hurts}
\label{sec:ana_pretrain}

\begin{table}[h]
\centering
\small
\resizebox{0.45\textwidth}{!}{
\begin{tabular}{lcc}
\toprule
\textbf{SAC ($K{=}34$, $15\times$)} & \textbf{ID F1} & \textbf{OOD F1} \\
\midrule
Matryoshka in SFT only        & \textbf{53.82} & \textbf{40.10} \\
Matryoshka in pretrain + SFT  & 51.51 & 37.60 \\
\bottomrule
\end{tabular}}
\caption{Applying budget sampling already during compressor pretraining degrades quality; FlexComp therefore applies it only in the SFT stage. 
}
\label{tab:pretrain}
\end{table}
A natural question is why we restrict Matryoshka training to the fine-tuning stage rather than incorporating it into pretraining. We tried the latter: sampling the budget in both stages. Table~\ref{tab:pretrain} shows the result on SAC at $K{=}34$: pretraining with mixed budgets loses $2.3$ ID and $2.5$ OOD F1 compared with sampling budgets only during SFT. Our interpretation is that pretraining is where the compressor learns the basic skill of packing text into memory tokens, and randomizing the slot count during this phase forces the model to split capacity across objectives before that skill is stable; once the base ability exists, adapting it to multiple budgets is cheap.
Practically, this is good news: Matryoshka training drops in at the cheapest stage of the pipeline.

\subsection{Balanced Labels Prevent Predictor Collapse}
\label{sec:ana_predictor}

\begin{table}[t]
\centering
\small
\setlength{\tabcolsep}{5pt}
\resizebox{0.45\textwidth}{!}{
\begin{tabular}{llccc}
\toprule
\textbf{Training set} & \textbf{Ep.} & \textbf{$K{=}1$} & \textbf{$K{=}10$} & \textbf{$K{=}34$} \\
\midrule
\multirow{2}{*}{Full ($3.38{:}1{:}1.18$)}
 & 1 & 89.98 & 13.33 & 36.77 \\
 & 2 & 90.12 & 15.61 & 36.81 \\
\midrule
\multirow{2}{*}{Resampled ($1{:}1{:}1$)}
 & 1 & 80.12 & 72.35 & 78.89 \\
 & 2 & 81.67 & 72.88 & 79.63 \\
\bottomrule
\end{tabular}}
\caption{Per-class accuracy of the $K$ predictor. Trained on the raw label distribution, the predictor collapses onto the majority class ($K{=}1$); balanced resampling restores usable accuracy on all three classes.}
\label{tab:predictor_acc}
\end{table}

Table~\ref{tab:predictor_acc} looks inside the $K$ predictor. Trained on the raw minimal-budget labels, whose distribution is skewed toward $K{=}1$ ($3.38{:}1{:}1.18$), the predictor takes the shortcut we warned about in \S\ref{sec:kpred}: it gets $90\%$ of the majority class right but only $13$--$16\%$ of the $K{=}10$ class, essentially defaulting to ``compress everything''. Resampling to $1{:}1{:}1$ trades a few points on the majority class for points on the minority class, and a second epoch adds little either way. We attribute this problem as the label distribution, not undertraining.




\subsection{Decoder Confidence Tracks Answer Quality}
\label{sec:confidence-analysis}
\begin{figure}
    \centering
    \includegraphics[width=0.9\linewidth]{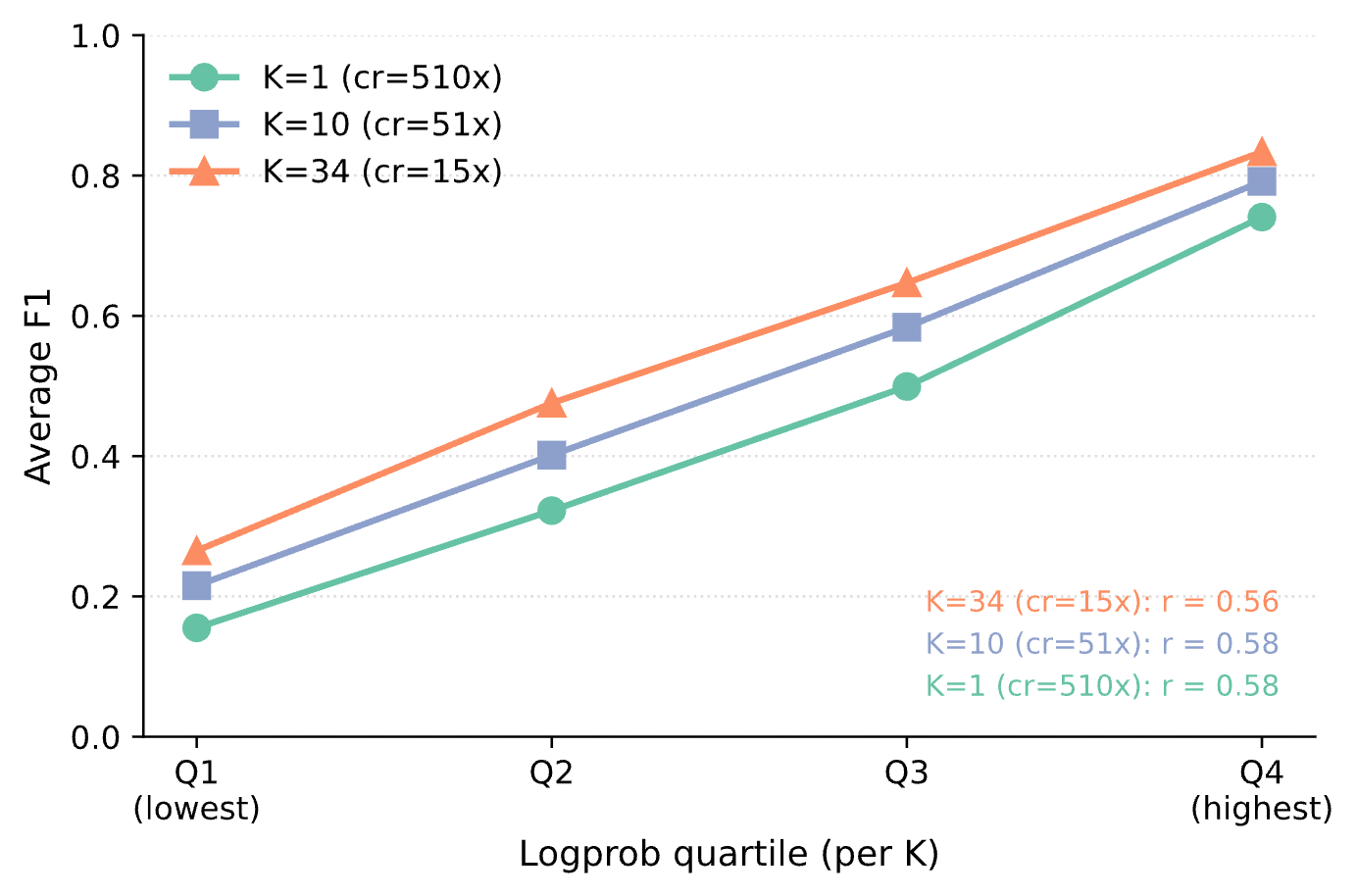}
    \caption{Spearman correlation between decoder confidence and answer
F1, computed on the Matryoshka-trained SAC model at each
$K$. Higher confidence consistently corresponds to
higher answer quality.}
    \label{fig:correlation}
\end{figure}
The cascade rests on the premise that the decoder's own confidence
identifies which answers to trust. Figure~\ref{fig:correlation}
validates this: across all budgets, the length-normalized answer
log-probability correlates with token-level F1 at Spearman's
$\rho = 0.56$ to $\rho = 0.58$, and the relationship holds at every
$K$, including $K{=}1$, where the signal must separate genuinely
solved inputs from confidently wrong ones. This is what makes the
calibrated thresholds of §\ref{sec:cascade} meaningful control
knobs rather than arbitrary cutoffs.

\subsection{From KV Memory to Serving Throughput}
\label{sec:throughput}
Table~\ref{tab:efficiency} reports what compression buys at inference
time on the Matryoshka-trained SAC model, at a batch size of 96 with
two context lengths. 

The memory story is direct: at both context lengths, moving from
15$\times$ to 510$\times$ shrinks the batch context KV cache by more
than 96\%. Because batched decoding is bandwidth-bound and model
weights are shared across the batch, per-sequence KV reads dominate
the incremental memory traffic, so this reduction converts into
aggregate decoding throughput. The effect grows with context length,
as the compressed context then accounts for a larger share of each
sequence's cache: The $K$~predictor inherits the
benefit while holding accuracy, cutting memory by 50\% and raising
throughput by 47\% at the longer context. The gains thus accrue
where compression is needed most.

\begin{table}[t]
\centering

\resizebox{0.48\textwidth}{!}{
\begin{tabular}{lcccc}
\toprule
\textbf{Ratio} & \textbf{KV Cache (MB)} & \textbf{Reduction}  & \textbf{Throughput (tok/s)} & \textbf{Speedup} \\

\midrule
\multicolumn{5}{c}{\emph{Context length 2{,}048}} \\
\midrule

15x  & 411 & -0\% & 5947 & +0\% \\
51x  & 123 & -70.1\% & 7657 & +28.9\% \\
510x & 15  & -96.4\% & 9531 & +60.3\% \\
K predictor & 230 & -44.0\%  & 7350 & +23.6\% \\

\midrule
\multicolumn{5}{c}{\emph{Context length 8{,}192}} \\
\midrule

15x  & 1638 & -0\% & 3920 & +0\% \\
51x  & 483 & -70.5\% & 6140 & +56.6\% \\
510x & 51  & -96.9\% & 7791 & +98.8\% \\
K predictor & 811 & -50.5\%  & 5764 & +47.0\% \\

\bottomrule
\end{tabular}}
\caption{Inference efficiency across compression ratios on the Matryoshka-trained SAC model
(generation length 512, batch size 96, greedy
decoding). \emph{KV Cache} is the total context KV memory over the
batch under grouped-query attention; \emph{Throughput} is the aggregate
decoding rate over the full batch.}
\label{tab:efficiency}
\end{table}

\subsection{More Memory Tokens Are Not Always Better}
\label{sec:analysis_nonmono}
\begin{figure}
    \centering
    \includegraphics[width=1.0\linewidth]{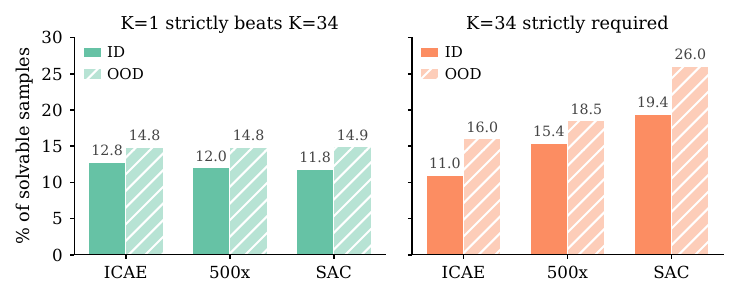}
    \caption{Budget non-monotonicity on solvable samples (samples
answered correctly (F1 $>$ 0) under at least one budget).
\textbf{Left:} fraction of samples where K{=}1 strictly outperforms
K{=}34. Larger budgets can actively hurt, and consistently more so
on OOD, where the gist-only representation avoids misleading surface details. \textbf{Right:} fraction of samples strictly requiring the
full budget (answered correctly only at K{=}34). Budget
sensitivity grows with compressor strength (ICAE $<$
500xCompressor $<$ SAC).}
    \label{fig:nonmono}
\end{figure}
Fixed-ratio evaluation reports averages, and averages hide a
phenomenon that per-sample analysis reveals: for a non-trivial
fraction of inputs, larger budgets actively \emph{hurt}. Throughout
this section we restrict analysis to \emph{solvable} samples (F1
$>$ 0 under at least one budget) and exclude the roughly 20\% to 30\% of samples that fail at every $K$.

Figure~\ref{fig:nonmono} (left) quantifies the effect at two levels
of severity. Across the three compressors, $K{=}1$ achieves strictly
higher F1 than $K{=}34$ on 11.8--12.8\% of solvable in-domain
samples; within these, on 6.8--7.7\% the model fails completely at
$K{=}34$ yet answers correctly from a single memory token, not a
degradation but a reversal. Figure~\ref{fig:nonmono} (right) adds a forward-looking observation:
the fraction of samples strictly requiring the full budget
(11.0--26.0\%) grows both with compressor strength
(ICAE $<$ 500xCompressor $<$ SAC). Additional
memory tokens pay off only when the compressor is capable enough to
exploit them. Adaptive budget allocation thus becomes \emph{more},
not less, important as compressors improve.
\begin{figure*}[t]
\centering
\begin{minipage}[t]{0.48\textwidth}
\begin{tcolorbox}[colback=gray!4, colframe=gray!40, boxrule=0.5pt,
                  left=4pt, right=4pt, top=3pt, bottom=3pt]
\small
\textbf{(a) Context (NaturalQuestions, ID):} ``Michael Casimir `Mike'
Stivic is a fictional character on the 1970s American television
sitcom All in the Family. [...] Michael was the husband of
Archie's daughter Gloria (played by \underline{Sally Struthers}).
\textbf{Rob Reiner} played the role of Michael Stivic throughout
the series.''

\vspace{2pt}
\textbf{Q:} Who played Mike Stivic on All in the Family?\quad
\textbf{Gold:} Rob Reiner

\vspace{1pt}
\begin{tabular}{@{}ll@{}}
K{=}1 (510$\times$): & Rob Reiner \hfill \textcolor{green!50!black}{\ding{51}} \\
K{=}34 (15$\times$): & Sally Struthers \hfill \textcolor{red!70!black}{\ding{55}} \\
\end{tabular}
\tcbline
\textbf{(b) Context (NaturalQuestions, ID):} ``Production was set to
begin in first quarter of 2014. Principal photography commenced on
March 6, 2014 in \textbf{Morocco}. Filming also took place in
Hurghada in Egypt, as well as in Berlin and D\"usseldorf in
Germany. Shooting wrapped in June 2014.''

\vspace{2pt}
\textbf{Q:} Where was A Hologram for the King filmed?\quad
\textbf{Gold:} Morocco / Egypt / Germany

\vspace{1pt}
\begin{tabular}{@{}ll@{}}
K{=}1 (510$\times$): & Morocco \hfill \textcolor{green!50!black}{\ding{51}} \\
K{=}34 (15$\times$): & Hollywood \hfill \textcolor{red!70!black}{\ding{55}} \\
\end{tabular}
\end{tcolorbox}
\end{minipage}\hfill
\begin{minipage}[t]{0.48\textwidth}
\begin{tcolorbox}[colback=gray!4, colframe=gray!40, boxrule=0.5pt,
                  left=4pt, right=4pt, top=3pt, bottom=3pt]
\small
\textbf{(c) Context (RACE, OOD, abridged):}
``\underline{Psychologists} in Britain have said that the last
full week of January is the most depressing time of year, and
labelled next Monday `\textbf{Blue Monday}'. [...] The bad weather
in January can also contribute to people feeling fed up. [...]''

\vspace{2pt}
\textbf{Q:} What's the best title of the passage? \\
\textbf{Gold:} Blue Monday

\vspace{1pt}
\begin{tabular}{@{}ll@{}}
K{=}1 (510$\times$): & Blue Monday \hfill \textcolor{green!50!black}{\ding{51}} \\
K{=}34 (15$\times$): & Psychologists \hfill \textcolor{red!70!black}{\ding{55}} \\
\end{tabular}
\tcbline
\textbf{(d) Context (RelationExtraction, OOD):} ``His Girl Friday is a
1940 American screwball comedy film directed by Howard Hawks, from
an adaptation by \underline{Charles Lederer, Ben Hecht and Charles
MacArthur} of the play \textbf{The Front Page} by Hecht and
MacArthur.''

\vspace{2pt}
\textbf{Q:} The His Girl Friday is based upon what? \\
\textbf{Gold:} The Front Page

\vspace{1pt}
\begin{tabular}{@{}ll@{}}
K{=}1 (510$\times$): & The Front Page \hfill \textcolor{green!50!black}{\ding{51}} \\
K{=}34 (15$\times$): & Charles MacArthur \& Ben Hecht \hfill \textcolor{red!70!black}{\ding{55}} \\
\end{tabular}
\end{tcolorbox}
\end{minipage}
\caption{Budget non-monotonic cases ($K{=}1$ correct, $K{=}34$
wrong) from the Matryoshka-trained SAC model, in domain (a,\,b) and out of domain (c,\,d); failure-mode
analysis is provided in Section~\ref{sec:analysis_nonmono}. Underlined: the in-context source of the
$K{=}34$ error; bold: the gold answer.}
\label{fig:qualitative}
\end{figure*}

\begin{table}[h]
\centering
\small
\setlength{\tabcolsep}{4pt}
\resizebox{0.45\textwidth}{!}{
\begin{tabular}{lcccc}
\toprule
\textbf{Config} & \textbf{ID F1} & \textbf{OOD F1} & \textbf{Avg.\ cr} & $\bar{K}$ \\
\midrule
\multicolumn{5}{c}{\emph{Fixed-ratio specialists (3 models)}} \\
\midrule
SAC 15$\times$   & 67.35 & 53.30 & 15$\times$  & 34 \\
SAC 51$\times$   & 61.44 & 46.89 & 51$\times$  & 10 \\
SAC 510$\times$  & 50.67 & 38.01 & 510$\times$ & 1  \\
\midrule
\multicolumn{5}{c}{\emph{Matryoshka (1 model, \textbf{ours})}} \\
\midrule
SAC 15$\times$   & \textbf{68.11} & \textbf{53.99} & 15$\times$  & 34 \\
SAC 51$\times$   & \textbf{61.97} & \textbf{47.36} & 51$\times$  & 10 \\
SAC 510$\times$  & \textbf{51.29} & \textbf{38.68} & 510$\times$ & 1  \\
\midrule
\multicolumn{5}{c}{\emph{Adaptive budget selection (\textbf{ours})}} \\
\midrule
Cascade (cons.)    & 67.92 & 53.96 & 156$\times$ & 25.6 \\
Cascade (bal.)     & 67.52 & 53.18 & 164$\times$ & 18.1 \\
Cascade (aggr.)    & 67.23 & 53.09 & 203$\times$ & 17.1 \\
Cascade (high cr)  & 66.81 & 52.95 & 241$\times$ & 16.1 \\
K predictor        & 67.69  & 53.27  &  183$\times$      & 14.2 \\ 
\bottomrule
\end{tabular}}
\caption{Scaling to Llama-3.1-8B (SAC; token-level F1, averaged over
ID / OOD tasks). The single Matryoshka model matches or exceeds its
fixed-ratio specialists at every budget, and adaptive selection substantially raises the  average compression ratio. $\bar{K}$ includes retry costs for cascade routing.}
\label{tab:8b}
\end{table}
We provide case studies in Figure~\ref{fig:qualitative}, which fall
into three failure modes. Panel~(a) shows \emph{relation confusion}:
the parenthetical ``Gloria (played by Sally Struthers)'' sits
directly beside the target relation, the 34-token memory preserves
both ``played by'' facts, and the decoder binds the question to the
co-occurring one. Panel~(b) is the most striking mode,
\emph{parametric-prior fallback}: the wrong answer
(``Hollywood'') appears nowhere in the context, faced with several
competing filming locations preserved in the 34-token memory, the
decoder abandons the memory altogether and falls back on its prior
for where films are made. Panel~(c) shows \emph{salient-entity
anchoring}: the passage opens with ``Psychologists,'' and the
34-token memory preserves this surface anchor, which the decoder
promotes to a title. Panel~(d) is the out-of-domain counterpart of
(a), with the confusion packed into a single sentence. In every case the larger
memory faithfully preserves surface details that create room for
error, whereas the single-token memory, forced to commit to the gist
of the context, does not. Extreme compression thus acts as an
information bottleneck: less capacity, more selectivity.

\subsection{Scaling to Llama-3.1-8B}
\label{sec:8b}

To test whether the picture transfers beyond 1B, we repeat the core
experiments with Llama-3.1-8B\footnote{\url{https://huggingface.co/meta-llama/Llama-3.1-8B}} as encoder and decoder on the SAC method. As shown in Table~\ref{tab:8b}, all three
findings transfer, and the first strengthens: the single Matryoshka
model now matches or exceeds its fixed-ratio specialists at
\emph{every} budget, in and out of domain (deltas of $+0.47$ to
$+0.76$ F1 across all six cells). Adaptive selection likewise carries over. The cascade's
conservative point reaches 156$\times$ average compression at
essentially no cost ($-0.19$ ID, $-0.03$ OOD F1 against the
Matryoshka 15$\times$ baseline), and its most aggressive point
241$\times$ within 1.3 ID F1. The $K$~predictor reaches 183$\times$
in a single compression-decoding pass at $-0.42$ ID and $-0.72$ OOD
F1 score, compared to the Matryoshka 15$\times$ baseline. With 183$\times$ compression it still matches the original fixed-ratio specialist run at 15$\times$ ($+0.34$ ID, $-0.03$ OOD).

\section{Conclusions}
FlexComp decouples the fixed compression ratio of soft context
compressors from both training and deployment: Matryoshka training
turns one model into an any-ratio compressor that matches the
accuracy of separately trained specialists, and per-input budget
selection (cascade routing or a single-pass $K$~predictor) preserves
the mildest ratio's accuracy at 158-266$\times$ average compression,
converting at serving scale into 50\% less context KV memory and
47\% higher decoding throughput. Our analysis reveals that larger
budgets can sometimes hurt, as extreme compression sheds misleading
surface details. Taken together, these results suggest that the
compression ratio need not be a design constant fixed once for all
inputs, but a per-input decision that a single model can both
support and make.


\bibliography{custom}

\appendix



\end{document}